\documentclass[default,iicol]{sn-jnl}

\usepackage{array}

\usepackage{graphicx}
\usepackage{multirow}
\usepackage{amsmath,amssymb,amsfonts}
\usepackage{amsthm}
\usepackage{mathrsfs}
\usepackage[title]{appendix}
\usepackage{xcolor}
\usepackage{textcomp}
\usepackage{manyfoot}
\usepackage{booktabs}
\usepackage{algorithm}
\usepackage{algorithmicx}
\usepackage{algpseudocode}
\usepackage{listings}
\usepackage{adjustbox}
\usepackage{caption}
\usepackage{float}
\usepackage{forest}
\usepackage{microtype}
\usepackage[T1]{fontenc}
\usepackage{multirow}
\usepackage{tabularx}
\usepackage{blindtext}

\usepackage[framemethod=TikZ]{mdframed}
\usepackage{listings}
\usepackage{colortbl}
\definecolor{codegreen}{rgb}{0,0.6,0}
\definecolor{codegray}{rgb}{0.5,0.5,0.5}
\definecolor{codepurple}{rgb}{0.58,0,0.82}
\definecolor{backcolour}{rgb}{0.95,0.95,0.92}

\lstdefinestyle{mystyle}{
    commentstyle=\color{codegreen},
    keywordstyle=\color{magenta},
    stringstyle=\color{codepurple},
    basicstyle=\ttfamily\footnotesize,
    breakatwhitespace=false,         
    breaklines=true,                 
    captionpos=b,                    
    keepspaces=true,                 
    showspaces=false,                
    showstringspaces=false,
    showtabs=false,                  
    tabsize=2
}

\usepackage{soul}
\usepackage{tcolorbox}
\usepackage{kotex} 
\definecolor{deeppink}{RGB}{255,20,147}
\definecolor{deepblue}{RGB}{22,83,126}
\definecolor{lightblue}{RGB}{173,216,230}
\definecolor{lightviolet}{RGB}{238,130,238}
\definecolor{lightred}{RGB}{255,182,193}
\definecolor{lightgreen}{RGB}{144,238,144}
\definecolor{lightyellow}{RGB}{255,255,224}
\definecolor{fluorescentpink}{RGB}{240,168,158} 
\definecolor{fluorescentgreen}{RGB}{160,219,142} 
\definecolor{fluorescentpurple}{RGB}{204,153,255} 
\definecolor{fluorescentorange}{RGB}{255,165,0} 
\definecolor{fluorescentyellow}{RGB}{239,244,131} 

\newtcolorbox{mybox}[1][lightblue]{ 
  colback=#1,
  colframe=#1,
  boxrule=0pt,
  arc=0pt,
  outer arc=0pt,
  boxsep=0pt,
  left=2pt,
  right=2pt,
  top=1pt,
  bottom=1pt
}

\usepackage{makecell}
\usepackage{natbib}
\usepackage{lmodern}

\theoremstyle{thmstyleone}

\theoremstyle{thmstyletwo}

\theoremstyle{thmstylethree}

\begin{document}

\title[A Survey on Integration of Large Language Models with Intelligent Robots]{A Survey on Integration of Large Language Models with Intelligent Robots}

\author{\fnm{Yeseung} \sur{Kim}}\email{yeseung.kim.88@kaist.ac.kr}

\author{\fnm{Dohyun} \sur{Kim}}\email{dohyun141@kaist.ac.kr}

\author{\fnm{Jieun} \sur{Choi}}\email{jichoi0101@kaist.ac.kr}

\author{\fnm{Jisang} \sur{Park}}\email{jisang.park@kaist.ac.kr}

\author{\fnm{Nayoung} \sur{Oh}}\email{lightsalt@kaist.ac.kr}

\author*{\fnm{*Daehyung} \sur{Park}}\email{daehyung@kaist.ac.kr}

\affil{\orgname{Korea Advanced Institute of Science and Technology}, \orgaddress{\state{Daejeon}, \country{Republic of Korea}}}

\abstract{
In recent years, the integration of large language models (LLMs) has revolutionized the field of robotics, enabling robots to communicate, understand, and reason with human-like proficiency. This paper explores the multifaceted impact of LLMs on robotics, addressing key challenges and opportunities for leveraging these models across various domains. By categorizing and analyzing LLM applications within core robotics elements---communication, perception, planning, and control---we aim to provide actionable insights for researchers seeking to integrate LLMs into their robotic systems. Our investigation focuses on LLMs developed post-GPT-3.5, primarily in text-based modalities while also considering multimodal approaches for perception and control. We offer comprehensive guidelines and examples for prompt engineering, facilitating beginners' access to LLM-based robotics solutions. Through tutorial-level examples and structured prompt construction, we illustrate how LLM-guided enhancements can be seamlessly integrated into robotics applications. This survey serves as a roadmap for researchers navigating the evolving landscape of LLM-driven robotics, offering a comprehensive overview and practical guidance for harnessing the power of language models in robotics development.
}


\maketitle

\section{Introduction}\label{sec_intro}

Over the last decade, we have witnessed remarkable progress in the field of robotics in applying language models (LMs). This progress includes not only human-like communication but also understanding and reasoning capabilities of robots, thereby significantly improving their effectiveness across various tasks, from household chores to industrial operations~\cite{nyga2018grounding, hu2022natural}. In the early stage of work, the success stemmed from statistical models analyzing and predicting words in linguistic expressions. These models enable robots to interpret human commands~\cite{patki2020language, roy2019leveraging}, understand contexts~\cite{barber2016multimodal, arkin2020multimodal}, represent the world~\cite{howard2022intelligence}, and interact with humans~\cite{tellex2020robots}, albeit with a limited depth of understanding. Then, the adoption of Transformer architecture with self-attention mechanisms~\cite{vaswani2017attention}, particularly pre-trained LMs such as BERT~\cite{devlin2018bert}, has elevated the capability of capturing complex patterns while fine-tuning models for specific tasks. However, the performance of these models is often contingent upon limited datasets, constraining their ability to grasp deeper contextual understanding and generalize across diverse scenarios.

With the advancement of large language models (LLMs), language-based robotics introduce innovative changes across various domains such as information retrieval, reasoning tasks, adaptation to environments, continuous learning and improvements, etc.~\cite{brohan2023can, huang2023inner}. These LLMs, characterized by their vast parameter sizes and training on internet-scale datasets, offer zero- and few-shot learning capabilities for downstream tasks without requiring additional parameter updates. These prominent advancements come from emergent abilities, defined as ``the abilities that are not present in small models but arise in large models'' in the literature~\cite{wei2022emergent}. The abilities have significantly enhanced robots' performance in understanding, inferring, and responding to open-set instructions by leveraging extensive common-sense knowledge~\cite{brown2020language}. Furthermore, prompt engineering techniques have enabled LLMs to incorporate richer contextual information through free-form language descriptions or interactive dialogues, facilitating generalized reasoning~\cite{wei2022chain}. 
The introduction of in-context learning abilities~\cite{brown2020language} leads LLMs to generate outputs in expected formats, such as JSON, YAML, or PDDL, or even code, based on provided instructions or demonstrations in prompts~\cite{guan2023leveraging, liang2023code}. Recent LLMs, such as GPT-4, have further expanded capabilities by integrating with external robotics tools such as planners or translators~\cite{liu2023llm+}. 

Despite the diverse capabilities of LLMs, their utilization faces several challenges~\cite{kaddour2023challenges}. Firstly, LLMs often generate inaccurate or unexpected responses. As the safety of robot execution is one of the most important deployment factors, LLM-based robotic applications require filtering and correction mechanisms to ensure safety. Second, the emergent abilities, such as in-context learning, are not predictable and consistent yet~\cite{chen2023relation}. Even minor alterations to input text may lead to unpredictable changes in response. 
Third, a well-designed prompt enables robots to effectively leverage the abilities of LLMs but there is a lack of systematic guidelines supporting key components of robotic systems, hindering seamless integration~\cite{firoozi2023foundation, zeng2023large, hu2023toward}. 
Therefore, we need to investigate component-wise LLM engagements in robotics toward an understanding of limitations and safety. 

Currently, various surveys have started exploring the intersection of LLMs and robotics~\cite{zeng2023large, vemprala2023chatgpt}, primarily focusing on application or interaction dimensions of LLM-based robotics. However, there remains a gap in providing holistic reviews and actionable insights for integrating LLMs across key elements of robotic systems, including communication, perception, planning, and control. Additionally, researchers explore the wide field of pre-trained large-capacity models, called foundation models, seeking the generalization capabilities across multimodal transformer-based models~\cite{hu2023toward, firoozi2023foundation}. However, this expansive field spans a wide spectrum of robotics and diverse methodologies, making emerging researchers miss in-depth reviews and guidelines. 

In this paper, as shown in Fig. \ref{fig:tree_all}, we aim to categorize and analyze how LLMs could enhance core elements of robotics systems and how we can guide emerging researchers in integrating LLMs within each domain, encompassing communication, perception, planning, and control, toward the development of intelligent robots. We structure this paper following three key questions: 
\begin{itemize}
    \item Q1: How are LLMs being utilized in each robotics domain?
    \item Q2: How can researchers overcome the integration limitation of LLMs?
    \item Q3: What basic prompt structures are required to produce a minimum functionality in each domain?
\end{itemize}
To address these questions, we focus on LLMs developed after the introduction of GPT-3.5~\cite{openai2023chatgpt}. We primarily consider text-based modalities but also review multimodalities for perception and control areas. However, for an in-depth review, we limit our investigation to LLMs rather than foundation models. 

In addition, we provide comprehensive guidelines and examples for prompt engineering, aimed at enabling beginners to access LLM-based robotics solutions. Our tutorial-level examples illustrate how fundamental functionalities of robotic components can be augmented or replaced by introducing four types of exemplary prompts: conversational prompt for interactive grounding, directive prompt for scene-graph generation, planning prompt for few-shot planning, and code-generation prompt for reward generation. By providing rules and tips for prompt construction, we outline the process of generating well-designed prompts to yield outputs in the desired format. These principles ensure effective LLM-guided enhancements in robotics applications, without parameter adjustments.

The remainder of this paper is organized as follows.
Section~\ref{sec_preliminary} outlines the historical background of LMs and LLMs in robotics. 
Section~\ref{sec_communication} reviews how LLMs empower robots to communicate via language understanding and generation.
Section \ref{sec_perception} investigate how LLMs perceive various sensor modalities and advance sensing behaviors.
Section \ref{sec_planning} and Section \ref{sec_control} organize LLM-based planning and control studies, respectively.
In Section \ref{sec_prompt}, we provide comprehensive guideline for prompt engineering as a starting point for LLM integration in robotics. 
Finally, Section \ref{sec_conc} summarizes this survey.

\noindent\begin{figure*}[ht!]
\begin{center}
    \centering
    \begin{adjustbox}{width=\textwidth} 
        \input{tree_all} 
    \end{adjustbox}
    \vspace{10pt}
    \caption{Overview structure of intelligent robotics research integrated with LLMs in this survey. The rightmost cells show the representative names (e.g., method, model, or authors) of papers in each category.}
    \label{fig:tree_all}
\end{center}
\end{figure*}
\noindent

\section{Preliminary}\label{sec_preliminary}
We briefly review language models used in robotics, categorizing them in terms of pre- and post-LLM eras.
Unlike previous literature~\cite{zeng2023large}, we define the pre-LLM era as the period for LMs up to the advent of GPT-2~\cite{radford2019language}, characterized by neural language models such as recurrent neural networks (RNNs)~\cite{elman1990finding} and early Transformer architectures. We then provide a brief explanation of LLMs, introducing terminologies and techniques for subsequent reviews.

\subsection{Language Models in Robotics}
In the pre-LLM era, early-stage studies have primarily focused on sequential data processing, using RNN-based models~\cite{hochreiter1997long, cho2014learning}. The models are often to transform linguistic commands into a sequence of actions~\cite{mei2016listen, blukis2018mapping} or formal languages~\cite{gopalan2018sequence}, leveraging RNN's sequence-to-sequence modeling capabilities. On the other hand, researchers have also used RNNs as language encoders converting textual input into linguistic features that could be mapped to visual features for referring object identification~\cite{roy2019leveraging, shridhar2020ingress}. However, the long-term dependency issue in RNNs restricts the scope of applications. Then, the Transformer architecture~\cite{vaswani2017attention}, which is a non-sequential model supporting long-range comprehension, has enabled new robotic tasks, such as vision-and-language navigation~\cite{chen2022think, chen2021topological}.

The studies in the pre-LLM era also show improved application performance unlike previous methods trained on small, task-specific datasets. Transformer-based models and self-supervised learning techniques, such as masked language modeling, have led to the development of internet-scale pre-trained models, including BERT~\cite{devlin2018bert} and GPT-2~\cite{radford2019language}.
These models exhibit a broad understanding of language, enabling both (1) improved generalization abilities and (2) fine-tuning for specific robotic tasks~\cite{kim2022natural, kim2023sggnet, shao2021concept2robot}.
In addition, researchers have also developed LMs that process multi-modal information~\cite{radford2021learning} since robotic applications often require accessing diverse multimodalities, such as natural language and vision, for interaction with users and environment~\cite{kim2024lingo, shridhar2022cliport}.

\subsection{Large Language Model in Robotics}
Recent advancements in LLMs, such as GPT-3~\cite{brown2020language}, GPT-4~\cite{achiam2024gpt}, LLaMA~\cite{touvron2023llama}, Llama 2~\cite{touvron2023llama2}, and Gemini~\cite{team2023gemini}, demonstrate notable improvements in understanding, contextual awareness, generalization capabilities, and knowledge richness, surpassing earlier language models. These improvements are from their training on vast datasets with billion-scale parameters, enabling them to capture intricate data patterns. Further, advanced learning strategies, such as reinforcement learning from human feedback, have been developed to align the behaviors of LLMs with human values or preferences~\cite{ouyang2022training}. However, learning with large-scale parameters requires computationally expensive costs of updating the entire model. To address this issue, researchers have developed parameter-efficient fine-tuning methods (e.g., an adapter~\cite{houlsby2019parameter} and LoRA~\cite{hu2022lora}) for robotic tasks. For example, \texttt{LLM-POP}~\cite{sun2023interactive} fine-tunes their model using adapters, which are small, trainable networks inserted into each layer of an LLM for interactive planning scenarios.

Alternatively, prompt engineering with in-context learning (ICL)~\cite{brown2020language} marks a significant advancement in learning from prompts without additional training. Its effectiveness relies on the design and quality of prompts, which can be enhanced with detailed task descriptions, few-shot examples, or model-friendly formats (e.g., `\#\#\#' as a stop symbol~\cite{zhao2023survey}). Moreover, chain-of-thought (CoT) prompting~\cite{wei2022chain} is another emerging approach that incorporates intermediate reasoning steps in prompts. 
The CoT method substantially enhances the reasoning and problem-solving capabilities of LLMs, making it a dominant technique in robotics applications~\cite{zeng2022socratic, liang2023code,singh2023progprompt}.

\section{Communication}\label{sec_communication}

We investigate the utilization of LLMs to facilitate human-like communication in robotics, enabling robots to interact with humans and other robotic agents effectively~\cite{mavridis2015review}. We categorize the communication capabilities into two primary areas: (1) language understanding and (2) language generation. We show the detailed categorization in Fig. \ref{fig:tree_all} alongside relevant studies, referred in green cells.

\subsection{Language Understanding}\label{subsec_understanding}
We review language understanding capabilities, addressing how LLMs handle the variability and ambiguity of linguistic inputs through \textit{interpretation} and \textit{grounding} processes.

\textit{Interpretation} transforms natural-language inputs into semantic representations that are easier for robots to process. These representations include formal languages such as linear temporal logic (LTL)~\cite{liu2023grounding, yang2023plug} and planning domain definition language (PDDL)~\cite{xie2023translating, liu2023llm+, chen2023autotamp, guan2023leveraging}, as well as programming languages such as Python~\cite{huang2023visual, kim2024lingo}.
To aid in interpreting free-form sentences, researchers leverage LLMs' ICL capabilities, providing guidelines and demonstrations within prompts~\cite{shah2023lm, huang2023visual, liu2023llm+, kim2024lingo}.
Despite the efforts, LLMs often fail to satisfy syntax or capture precise semantics when translating an input into formal languages. 
To address this issue, researchers suggest simplifying vocabulary or fine-tuning LLMs with domain-agnostic data~\cite{liu2023grounding, yang2023plug}.
For example, \texttt{Lang2LTL}~\cite{liu2023lang2ltl} translates landmark-referring expressions in navigational commands into LTL symbols.
Further improvements often involve in using human feedback and syntax checkers to correct generated formal language translations~\cite{chen2023autotamp, guan2023leveraging}.
For instance, Guan et al. present a human-in-the-loop translation framework, in which human domain experts repeatedly review PDDL descriptions and provide feedback in natural language~\cite{guan2023leveraging}.

\textit{Grounding} is another process that maps linguistic expressions to reference targets, such as behaviors or objects, recognizable to robots. Early studies identify mappings that maximize the cosine similarity between word embeddings of LLM outputs and real-world targets~\cite{huang2022language, raman2023cape, liu2023grounding, kim2024lingo}.
Subsequent studies leverage LLMs' commonsense knowledge to capture the context of object text labels for improved grounding~\cite{rana2023sayplan, gu2023conceptgraphs}. For instance, \texttt{ConceptGraphs}~\cite{gu2023conceptgraphs} demonstrate how LLMs can ground an expression, `something to use as a paperweight,' to a ceramic vase based on size and weight assumptions.
However, grounding accuracy depends on the detail and accuracy of the world model. To address this, researchers augment LLMs with multimodal capabilities to directly correlate linguistic inputs with sensory percepts~\cite{driess2023palm, yang2023lidar, hong20233d, qian2024affordancellm}, or enable LLMs to interact with environments~\cite{yang2023llm, zhao2023chat} or humans~\cite{huang2023inner, ren2023robots, park2024clara} for better context gathering.
For instance, \texttt{LLM-Grounder}~\cite{yang2023llm}, a 3D visual grounding method, actively gathers environmental information using vision tools such as LERF~\cite{kerr2023lerf} and OpenScene~\cite{peng2023openscene}.

\subsection{Language Generation}\label{subsec_generation}
Language generation refers to the production of human-like written or spoken language that reflects communicative intents~\cite{gatt2018survey}. We categorize language generation into \textit{task-dependent} and \textit{-independent} types based on their communication intents, diverging from conventional natural language generation (NLG) categories of \textit{text-to-text} and \textit{data-to-text}~\cite{Dong2021ASO} due to our focus on the communicative purposes of studies.

\textit{Task-dependent} language generation focuses on producing language with specific functional objectives, being declarative or imperative. To generate open-ended declarative statements, researchers often provide LLMs with contextual information~\cite{chen2024scalable, mandi2023roco, hunt2024conversational}. 
However, LLMs often result in repetitive and factually inconsistent outputs, confined by the reliance on previous dialogues and commonsense knowledge \cite{chen2024scalable, li2020don}.
Consequently, researchers augment LLMs with auxiliary knowledge sources to broaden the scope of available information~\cite{axelsson2023you, yamazaki2023building, cherakara2023furchat}. For instance, Axelsson and Skantze enhance a robot museum guide with knowledge graphs \cite{axelsson2023you}. Furthermore, researchers instruct LLMs to clarify ambiguities by generating imperative instructions requesting human assistance \cite{huang2023inner, dai2023think}. To improve inference steps, probabilistic models have been introduced to evaluate the uncertainty of situations~\cite{ren2023robots, park2024clara}. For instance, \texttt{KnowNo}~\cite{ren2023robots} and \texttt{CLARA}~\cite{park2024clara} interaction systems assess confidence and semantic variance, respectively, triggering generation only when these metrics indicate significant uncertainty.

\textit{Task-independent} language generation involves crafting expressions with social-emotional objectives~\cite{chattaraman2019should}, by embedding non-verbal cues (e.g., non-verbal sounds, hand gestures, and facial expressions) within prompts to enhance engagement and empathy~\cite{khoo2023sll, lee2023developing}. For example, Khoo et al. have developed a conversational robot that generates empathetic responses using transcribed audio and visual cues \cite{khoo2023sll}. However, conversations with LLMs remain superficial due to the limited knowledge and dialogue history~\cite{irfan2023between}. To overcome this, researchers integrate memory modules into LLMs, enabling them to distill and store information from conversations in a structured format~\cite{irfan2023between, ichikura2023method, cho2023story, yu2024affordable}. For example, the companion robot designed by Irfan et al. continuously updates the robot's memory based on interactions with users to generate personalized dialogues \cite{irfan2023between}.

\section{Perception}\label{sec_perception}

Perception plays a crucial role in enabling robots to make decisions, plan their actions, and navigate the real world \cite{premebida2018intelligent}.
In the field of LLM-based robotic perception, research primarily focuses on two aspects: sensing modalities and behaviors.
In this section, we introduce how LLM-based robots have integrated language with sensor modalities and how agents acquire environmental information through passive and active perception behaviors. 
Fig. \ref{fig:tree_all} presents the detailed categorization alongside relevant studies, referred in pink cells.

\subsection{Sensing Modalities}\label{subsec_modality}
Researchers have significantly advanced robots' comprehension and generalization capabilities through the integration of multimodal language models. We categorize primary sensing modalities into \textit{visual}, \textit{auditory}, and \textit{haptic} modalities, reviewing recent studies leveraging multimodal LLMs for perception tasks.

\textit{Visual} perception tasks involve the interpretation of visual information such as image or point clouds. Pre-trained visual-language models (VLMs), such as CLIP \cite{radford2021learning} and InstructBLIP  \cite{li2023blip}, allows LLM-based robots to directly utilize image sources. 
For instance, recent LLM-based manipulation systems, such as \texttt{TidyBot}~\cite{wu2023tidybot} and \texttt{RoCo}~\cite{mandi2023roco}, use image-inferred object labels or scene descriptions generated from CLIP and OWL-ViT~\cite{minderer2022simple}, respectively. 
In addition, researchers extend reasoning capabilities by applying VLMs on downstream tasks such as image captioning \cite{gu2023conceptgraphs} and visual question answering (VQA) \cite{gao2024physically,kwon2024toward, mirjalili2023lan}. The downstream tasks enable LLMs to subsequently request VLMs to infer object properties (e.g., material, fragility)~\cite{gao2024physically} or ground object parts for grasping~\cite{mirjalili2023lan}. However, images are often challenging to acquire spatial-geometric information. 

Alternatively, Huang et al. associate visual-language features from a VLM (i.e., LSeg~\cite{li2022language}) with three-dimensional (3D) point clouds for 3D map reconstruction \cite{huang2023visual}. Further, Jatavallabhula et al. improve this association mechanism with RGB-D images by introducing fine-grained and pixel-aligned features from VLMs \cite{jatavallabhula2023conceptfusion}. However, association with 3D information tends to be memory intensive, limiting scalability for large scenes \cite{huang2023visual, jatavallabhula2023conceptfusion, yang2023llm}. As an alternative solution, researchers often associate geometric and semantic features with 3D scene graphs~\cite{gu2023conceptgraphs}. 

\textit{Auditory} perception involves the interpretation of sound. LLM-based studies often leverage pre-trained audio-language models (ALMs), such as AudioCLIP \cite{guzhov2022audioclip} and Wav2CLIP \cite{wu2022wav2clip}, integrating them with visual data to enhance environmental or contextual understanding \cite{huang2023audio, liu2023reflect, zeng2022socratic, shah2023mutex}. For example, \texttt{AVLMaps} \cite{huang2023audio}, a 3D spatial map constructor with cross-modal information, integrates audio, visual, and language signals into 3D maps, enabling agents to navigate using multimodal objectives such as ``move between the image of a refrigerator and the sound of breaking glass.'' In addition, \texttt{REFLECT} \cite{liu2023reflect}, a framework for summarizing robot failures, transforms multi-sensory observations such as RGB-D images, audio clips, and robot states into textual descriptions to enhance LLM-based failure reasoning.

\textit{Haptic} perception involves the interpretation of contact information. Researchers introduce multimodal perception modules that interactively incorporate \textit{haptic} features obtained from pre-defined high-level descriptions~\cite{zhao2023chat} or CLIP-based tactile-image features~\cite{hong2024multiply} about haptic interactions. For example, \texttt{MultiPLY}~\cite{hong2024multiply}, a multisensory LLM, converts tactile sensory readings into a heatmap, encoded by CLIP. Then, by introducing a linear layer of tactile projector, the model maps the heatmap information to the feature space of LLMs.

\subsection{Sensing Behavior}\label{subsec_activity} 
Following the type of sensing behaviors, we decompose this section into \textit{passive} and \textit{active} perceptions. 

The \textit{passive} perception refers to the process of gathering sensory information without actively seeking it out. Despite its limited nature, \textit{passive} sensing has been extensively employed in LLM-based robotics studies for various tasks: object recognition \cite{gao2024physically, wu2023tidybot, hu2023look}, pose estimation \cite{mirjalili2023lan, xu2023reasoning}, scene reconstruction \cite{gu2023conceptgraphs, huang2023voxposer, shah2023lm, shah2023lm}, and object grounding \cite{jatavallabhula2023conceptfusion, yang2023llm, wang2023wall}.  For example, \texttt{TidyBot} \cite{wu2023tidybot} detects the closest object from an overhead view and subsequently recognizes its object category using a closer view captured by the robot's camera. However, the \textit{passive} nature of sensing limits the ability to perform tasks when information is unobserved or unavailable (e.g., unseen area, weight).

On the other hand, \textit{active} perception refers to the conscious process of gathering sensory information by taking additional actions. \textit{Active} information gathering enhances environmental understanding by acquiring new information through sensory observations or requesting user feedback~\cite{song2023llm, kwon2024toward}. For example, \texttt{LLM-Planner} \cite{song2023llm} generates seeking actions such as `open the refrigerator' to locate invisible objects. Recent studies also focus on collecting sensory data to better understand objects' physical properties~\cite{zhao2023chat, hong2024multiply, sun2023interactive}. However, LLMs often generate inaccurate or fabricated information, known as hallucinations. To address this issue, Dai et al. introduce a personalized conversational agent designed to ask users for uncertain information \cite{dai2023think}.

\section{Planning}\label{sec_planning}

Planning involves organizing actions to solve given problems, typically through generating a sequence of high-level symbolic operators (i.e., task planning) followed by executing them using low-level motor controllers~\cite{garrett2020pddlstream, li2021reactive}. This section investigates how LLM-based planning research addresses limitations in the planning domain by categorizing them into three key research areas: (1) task planning, (2) motion planning, and (3) task and motion planning (TAMP). Fig. \ref{fig:tree_all} presents the detailed categorization along with related planning studies, referred in purple cells.

\subsection{Task Planning}
LLM-based task planners are capable of generating plans without strict symbol definitions~\cite{huang2022language}, while traditional task planners are required to pre-define operators with domain knowledge about available actions and constraints~\cite{mcdermott1998pddl, FIKES1971189}. In this field, most planners employ a \textit{static} planning strategy, which takes fixed descriptions that are not adaptable to the changes in environment~\cite{zeng2022socratic}. However, an alternative approach, \textit{adaptive} planning, allows for the incorporation of environmental feedback into input prompts, enabling adjustments to actions based on observed conditions. This section reviews LLM-based planners in terms of these two strategies: \textit{static} and \textit{adaptive} planning. 

\noindent\textbf{Static planning}: \textit{Static planning} approaches are generally zero- or few-shot prediction methods, where zero-shot methods generate a plan based solely on an input command, while few-shot methods leverage learning from a limited set of similar examples \cite{cao2023robot, di2023towards, kannan2024smart, zeng2022socratic}. However, LLMs often exhibit poor performance in long-horizon task planning due to limited reasoning ability \cite{liu2023llm+, valmeekam2023planning}. To address this limitation, Huang et al. introduce a planner that iteratively selects the most probable action among executable ones generated by LLMs \cite{huang2022language}. 

Alternatively, LLM-based code generators, such as \texttt{Code as Policies}~\cite{liang2023code} or \texttt{ProgPrompt}~\cite{singh2023progprompt}, produce codes that result in actions responsive to observations~\cite{huang2023instruct2act, huang2023visual}. Singh et al. demonstrate that code generation outperforms basic task planning from LLMs since the output plan closely aligns with execution environments \cite{singh2023progprompt}. Despite their advantages, these methods lack validation and replanning processes. 

To validate plans, researchers often augment LLMs with logical programs, either to (1) check if resulting plans violate logical constraints or (2) generate plans using an external logical planner. For instance, \texttt{SayPlan} \cite{rana2023sayplan}, a GPT4-based planner, validates abstract-level actions through a scene-graph simulator 3DSG~\cite{agia2022taskography}, while \texttt{LLM+P} \cite{liu2023llm+} applies a PDDL problem translated from LLMs to a classical task planner, Fast Downward \cite{helmert2006fast}. In addition, Silver et al. demonstrate that a search-based planner with an initial plan from LLMs performs better by exploring fewer nodes \cite{silver2022pddl}. These studies underscore the effectiveness of integrating LLMs with logical programs to increase the success rate or the performance of generating feasible plans. 

\noindent\textbf{Adaptive planning}: \textit{Adaptive planning} allows robots to modify their plans or actions in response to feedback, either by generating new plans based on environmental observations~\cite{chen2024scalable, wu2023tidybot, vemprala2023chatgpt, zhao2023chat, zhao2023large} or by detecting failures and adjusting accordingly~\cite{huang2023inner}. Chen et al. and Huang et al. introduce adaptation strategies that generate new plans based on observed feedback, enabling robots to respond to a broader range of scenarios \cite{chen2023open, huang2023grounded}.

Another adaptation strategy is the detection of failures as feedback. For instance, \texttt{Inner Monologue}~\cite{huang2023inner} retries the initial plan until it succeeds. Furthermore, other studies provide textual explanations about past failures to help avoid recurrent issues~\cite{wang2023describe, lin2023swiftsage, liu2023reflect, raman2023cape}. \texttt{LLM-Planner} \cite{song2023llm} and \texttt{COWP} \cite{ding2023integrating} improve replanning capabilities by finding alternative plans that leverage context from observations and the commonsense knowledge of LLMs. These flexibilities in adapting to new information enhance robot autonomy in dynamic settings.

\subsection{Task and Motion Planning}
We outline LLM-based low-level planning, classifying methodologies into motion planning and TAMP areas. 

\textit{Motion planning} refers to the process of generating a path by computing sequential waypoints within configuration or task spaces. Jiao et al. introduce an LLM-based motion planner that directly generates positional sequences for drone choreography~\cite{jiao2023swarm}. While this work demonstrates LLMs' spatial reasoning abilities, the scenario presented is relatively simple. Further, planning spaces are often continuous, which presents a challenge for language models that operate with discrete tokens. Alternatively, indirect sequencing approaches, such as \texttt{VoxPoser}~\cite{huang2023voxposer}, generate a potential field code with the help of a VLM and then conduct motion planning within the generated field, augmenting LLMs with a search-based planner.

\textit{TAMP} refers to integrating high-level task planning with low-level motion planning. Recent studies often use LLMs as TAMP planners, leveraging LLMs' logical and physical reasoning capabilities~\cite{mandi2023roco, kwon2023language, xia2024kinematic}. Researchers guide LLMs to generate high-level subgoals, which are then used for low-level trajectory generation~\cite{mandi2023roco, kwon2023language}. However, the coarse representations of LLMs restrict their applications to simple tasks such as pick-and-place. To address this limitation, researchers use additional prompts or augment LLMs to improve reasoning abilities. For example, Xia et al. enable LLMs to consider kinematic knowledge through kinematic-aware prompting for more complex manipulation tasks, such as articulated object manipulation \cite{xia2024kinematic}. Ding et al. introduce a logic-augmented LLM planner that checks the logical feasibility of the task plans generated by LLMs \cite{ding2023task}. Meanwhile, others use physics-augmented LLM planners to evaluate physical feasibility~\cite{lin2023text2motion, chen2023autotamp, ha2023scaling}. For example, \texttt{Text2Motion}~\cite{lin2023text2motion} allows an LLM to generate physically feasible high-level actions and combine them with learned skills for low-level actions.

\section{Control}\label{sec_control}
Early studies primarily focus on establishing mappings between simple linguistic commands and known motion primitives.
With the advent of deep learning, researchers have explored two main approaches in control: direct modeling of control values based on linguistic instructions \cite{reed2022generalist, brohan2023rt} and indirect interpretation of complex instructions via LLMs to generate actions~\cite{xie2023text2reward}.
We categorize the work in this field into two groups: (1) direct approach which means the direct generation of control commands based on linguistic instructions and (2) indirect approach which stands for indirect specification of control commands through linguistic guidance.
Fig. \ref{fig:tree_all} presents a detailed categorization alongside related papers, referred in orange cells.

\subsection{Direct Approach}
The direct approach involves using an LLM to interpret and produce executable commands, either by selecting motion primitives \cite{tang2023saytap} or generating control signals \cite{zitkovich2023rt, wang2023prompt}.
Early work generates action tokens to produce a control policy training Transformer architecture with task-specific expert demonstrations such as \texttt{Gato}~\cite{reed2022generalist}, \texttt{RT-1}~\cite{brohan2023rt}, and \texttt{MOO}~\cite{stone2023open}. Researchers linearly map action tokens to discretized end-effector velocities \cite{reed2022generalist} or displacements \cite{brohan2023rt, stone2023open} for continuous motion.
While these approaches demonstrate a degree of generalization over unseen tasks, such as new objects or realistic instructions, they often require extensive data collection and training time.

To reduce the collection effort, researchers often leverage existing web-scale vision and language datasets such as \texttt{RT-2}~\cite{zitkovich2023rt} and \texttt{RT-X}~\cite{padalkar2023open}.
For example, Zitkovich et al. train VLMs (e.g., \texttt{PaLI-X} \cite{chen2023pali} and \texttt{PaLM-E} \cite{driess2023palm}) with visual-language datasets and robotic demonstrations \cite{zitkovich2023rt}. This approach maintains general knowledge of visual-language tasks while training for control tasks.
In addition, to reduce the training burden, Chen et al. use a low-rank adaptation (LoRA) \cite{hu2022lora} method for fine-tuning an LLM for control tasks rather than fine-tuning the entire model \cite{chen2023driving}.

LLMs often struggle to generate continuous action-level commands such as joint position and torque values, as LLMs typically generate atomic elements known as tokens \cite{tang2023saytap}. Therefore, researchers instead generate task-level output using LLMs \cite{tang2023saytap, mirchandani2023large, cao2023ground}. 
For example, \texttt{SayTap}, an LLM-based controller for walking, generates contact patterns between feet and the ground for walking motion with an LLM instead of directly producing joint positions~\cite{tang2023saytap}. Other studies solve the control problem as natural-language generation tasks, completing a sequence of end-effector poses~\cite{mirchandani2023large} or generating Python codes~\cite{cao2023ground}.
Recent studies often restrict action space to enhance LLM control outputs. 
For instance, Wang et al. design a prompt that produces positive integer control values while maintaining a smoothness trend in outputs \cite{wang2023prompt}.
Alternatively, Li et al. demonstrate that incorporating robot kinematics information helps the LLM determine joint values for desired poses \cite{li2023learning}.

\newcolumntype{C}[1]{>{\arraybackslash}m{#1\linewidth}}

\noindent\begin{figure*}[ht!]
    \centering
    \includegraphics[width=0.95\textwidth]{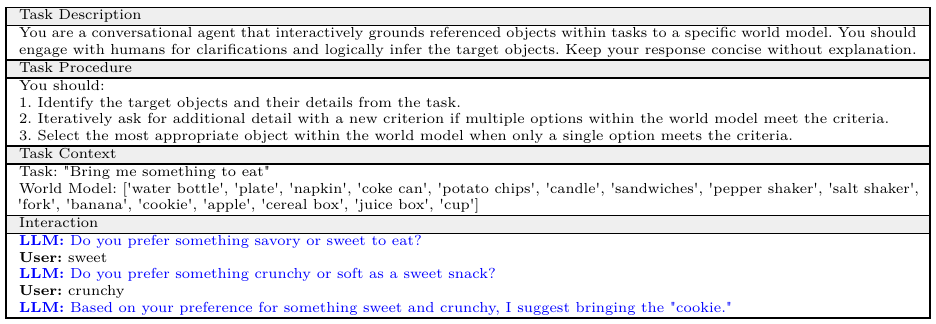}
    \caption{A conversational prompt for interactive grounding. 
Through the `task' in the prompt, we ask an LLM to ground the underspecified object, referred to as `something' in the `task' as a `cookie' by interactively asking for a personal preference.
The prompt consists of \textit{task description}, \textit{task procedure}, and \textit{task context} parts guiding the LLM's behavior and contextual understanding. 
The words in bold indicate the subjects of interactions, while the LLM responses are highlighted in blue.}
    \label{tab:prompt_NLP}
\end{figure*}
\noindent

\subsection{Indirect Approach}\label{sec_indirect_control}
LLMs are also useful for generating indirect representations of control commands, such as subgoals or reward functions, based on natural language instructions. To guide the learning process, researchers leverage the goal description that explains desired behaviors in natural language  \cite{du2023guiding, jia2023chain, kumar2023words}.
For example, \texttt{ELLM}~\cite{du2023guiding}, an LLM-based reinforcement learning (RL) framework, uses an LLM to generate subgoal descriptions as prior knowledge of the RL policy and further uses the similarity between current observation and the subgoal description in text embedding space to calculate reward. 
Further, Kumar et al. generate a goal description based on the history of human instructions to reuse previously learned skills \cite{kumar2023words}.
However, as the output of an LLM is a natural-language description, these approaches require an additional step of grounding or interpreting the description.

Alternatively, researchers often generate code-level reward functions.
Yu et al. convert a natural-language goal into a high-level motion description and then generate a corresponding reward function \cite{yu2023language}.
However, this method requires pre-defined reward formats. Instead, recent work prompts an LLM to infer a new reward function from human-designed examples \cite{katara2023gen2sim, wang2023robogen}.
Nonetheless, the generated reward functions may not always be accurate or optimal enough for direct use in training \cite{song2023self}.

To improve accuracy, researchers add a refinement loop to validate both the syntax \cite{perez2023larg} and semantics \cite{xie2023text2reward, song2023self, ma2023eureka, zeng2023learning} of the generated reward functions.
For example, Song et al. use an LLM to redesign a reward function based on the convergence of the training process and the resulting robot motion \cite{song2023self}. Chu et al. employ an LLM to directly generate rewards for evaluating robot motion \cite{chu2024accelerating}. Other approaches refine a motion by adjusting control parameters based on the error state~\cite{tagliabue2023real} or by selecting a suitable motion target from human feedback~\cite{liu2023interactive}.

\section{Prompt Guideline}\label{sec_prompt}
We provide prompt design guidelines for robotic tasks to researchers entering this field. A prompt is a message that directs LLMs to process and generate outputs according to our instructions~\cite{liu2023pre, white2023prompt}.
Well-designed prompts 
\begin{itemize}
\item include clear, concise, and specific statements without using technical jargon,
\item incorporate examples that allow anticipating the model's process,
\item specify the format that we want the output to be presented, and
\item contain instructions to constrain actions.
\end{itemize}
The prompts enable models to generate desired content following output formats and constraints without parameter updates. We provide guidelines over four robotics fields: (1) interactive grounding, (2) scene-graph generation, (3) few-shot planning, and (4) reward function generation. 

\subsection{Conversation Prompt: Interactive Grounding}\label{ssec_prompt_ig}
We detail a conversation prompt design, leveraging an LLM as a grounding agent, to clarify commands such as `Bring me something to eat' and infer the ambiguous targets, expressed as `something,' through logical inference. Fig.~\ref{tab:prompt_NLP} shows the design detail, where the prompt consists of three key components: \textit{task description}, \textit{task procedure}, and \textit{task context}. We further describe each component as follows.

The \textit{task description} outlines the expected behavior and response format of the LLM. In this example, we particularly emphasize its role as a conversational agent, which fosters dynamic interactions with users, guided by directives such as `you should.' Further, the imperative statements containing `keep' provide task constraints or requirements. We also place behavioral constraints at the end to suppress the LLM's verbosity. 

The \textit{task procedure} then defines a sequence of inference steps for the LLM to follow, aimed at achieving the task objective. This description employs numbered steps to instruct LLMs to execute the actions step by step. By using logical representations, we also enforce actions to be performed in a logical order; we use `iteratively' to indicate a `while loop' and `if' or `when' to represent conditions.

The \textit{task context} describes the contextual inputs, such as `world model,' that LLMs perform grounding upon. Consistency in terminology across \textit{task description} and \textit{task procedure} is crucial for LLM operations. For example, the common expressions, such as `task' and `world model,' allow the LLM to work within the same context provided. Further, by using clear names for objects in the world model, we enable the LLM to apply common knowledge to named entities. Note that although we use a list of objects as a world model, LLMs accept world models in various formats: textual descriptions, a list of objects, and scene graphs. 

With these structured components, the prompt invokes an interactive grounding dialogue for precise object identification, as shown in Fig.~\ref{tab:prompt_NLP}.
We obtain the resulting interaction using ChatGPT 3.5~\cite{openai2023chatgpt}.

\noindent\begin{figure*}[ht!]
    \centering
    \includegraphics[width=0.95\textwidth]{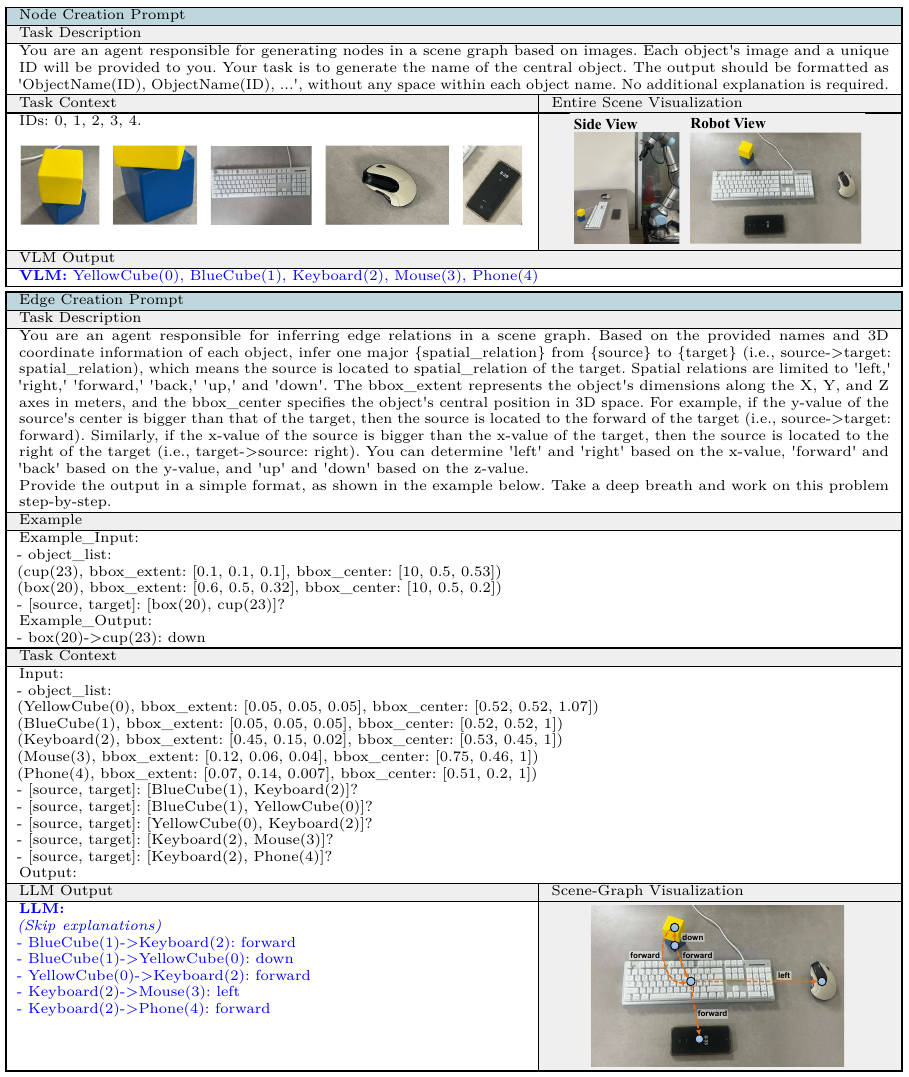}
    \caption{Directive Prompts for generating a scene graph. The table includes two prompts: \textit{node creation} and \textit{edge creation}. Given segmented images in a scene, a VLM identifies objects, and then an LLM infers relevant relationships using geometric information. The words in bold indicate the subjects of outputs, while the model responses are highlighted in blue. The visualizations are not part of any conversations.}
    \label{tab:prompt_perception_C}
\end{figure*}
\noindent

\subsection{Directive Prompt: Scene-Graph Generation}
We introduce directive prompt designs for constructing a scene graph from a scene image using a multimodal LLM, particularly with GPT-4~\cite{achiam2024gpt}. The scene graph consists of objects as nodes and their relationships as edges~\cite{fisher2011characterizing, gu2023conceptgraphs}. Despite the advancement of multimodal LLMs, their capability has limitations in inferring 3D relationships from a 2D image \cite{chen2024spatialvlm}. To reduce the limitation, we decompose the task into two steps: \textit{node creation} with multimodal inputs and \textit{edge creation} with textual information. We describe each step with detailed examples provided in Fig.~\ref{tab:prompt_perception_C}.

\noindent\begin{figure*}[ht!]
    \centering
    \includegraphics[width=0.95\textwidth]{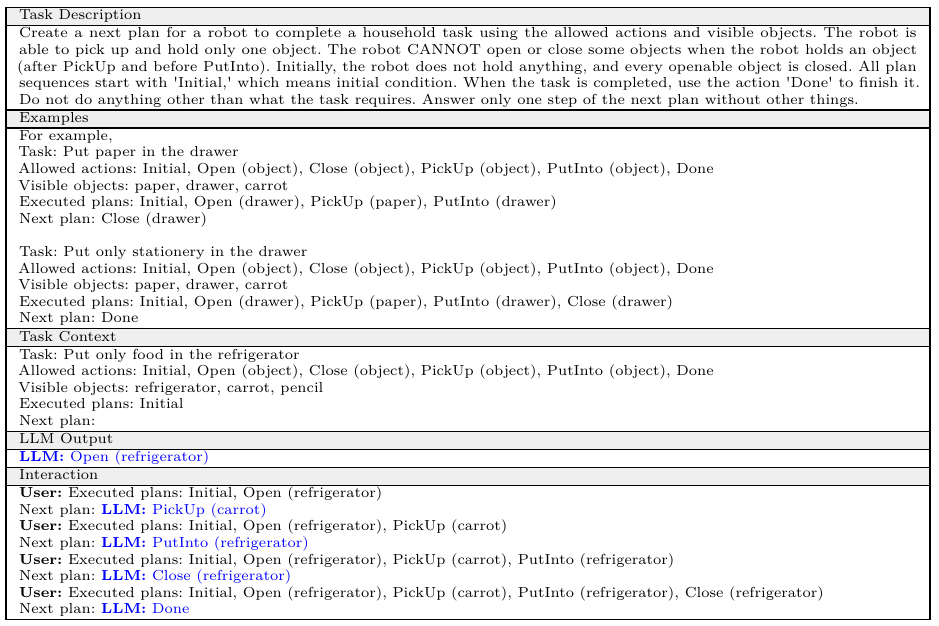}
    \caption{A planning prompt for few-shot planning. Leveraging input-output example pairs, the LLM improves performance in generating a plan to accomplish the task objective. The prompt consists of \textit{task description}, \textit{examples}, and \textit{task context}. The words in bold indicate the subjects of interactions, while the LLM responses are highlighted in blue.}
    \label{tab:prompt_planning}
\end{figure*}
\noindent

The prompt for \textit{node creation} consists of two parts: (1) \textit{task description} and (2) \textit{task context}. The \textit{task description} includes the multimodal LLM's expected behavior (i.e., role) and response format, similar to Section~\ref{ssec_prompt_ig}. For instance, the multimodal LLM's role is to identify objects as nodes in the given images. We then specify the output format as `ObjectName(ID)' for consistency and simplicity. Then, the \textit{task context} presents a sequence of unique object identifiers with corresponding object-centric images. 
In this scenario, we assume that object-centric images are obtained using an active perception method to identify objects with occlusion.

The \textit{edge creation} consists of (1) \textit{task description}, (2) \textit{examples}, and (3) \textit{task context}. The \textit{task description} not only specifies the expected behavior and output format but also elucidates on how to identify the relationships between nodes leveraging examples. We particularly explain how the LLM uses 3D object coordinates and unit measurements to infer spatial relationships from a predefined set such as `left,' `right,' etc. Unlike the node creation, this allows for generating additional output explanations to accommodate the complexity of discerning spatial relationships. 

To enhance the understanding of the input format and corresponding output, we include \textit{examples} showcasing edge generation. We choose an example similar to the target scenario in terms of objects and their spatial interrelationships, thereby providing richer information for edge identification. 
Finally, the \textit{task context} provides source and target node information as inputs and leaves the output empty to obtain responses from the LLM.
We also assume that the 3D bounding boxes are obtained through a neural network-based detector~\cite{Mousavian_2017_CVPR} or from the extent of point clouds using depth information~\cite{gu2023conceptgraphs}.

\noindent\begin{figure*}[ht!]
    \centering
    \includegraphics[width=0.95\textwidth]{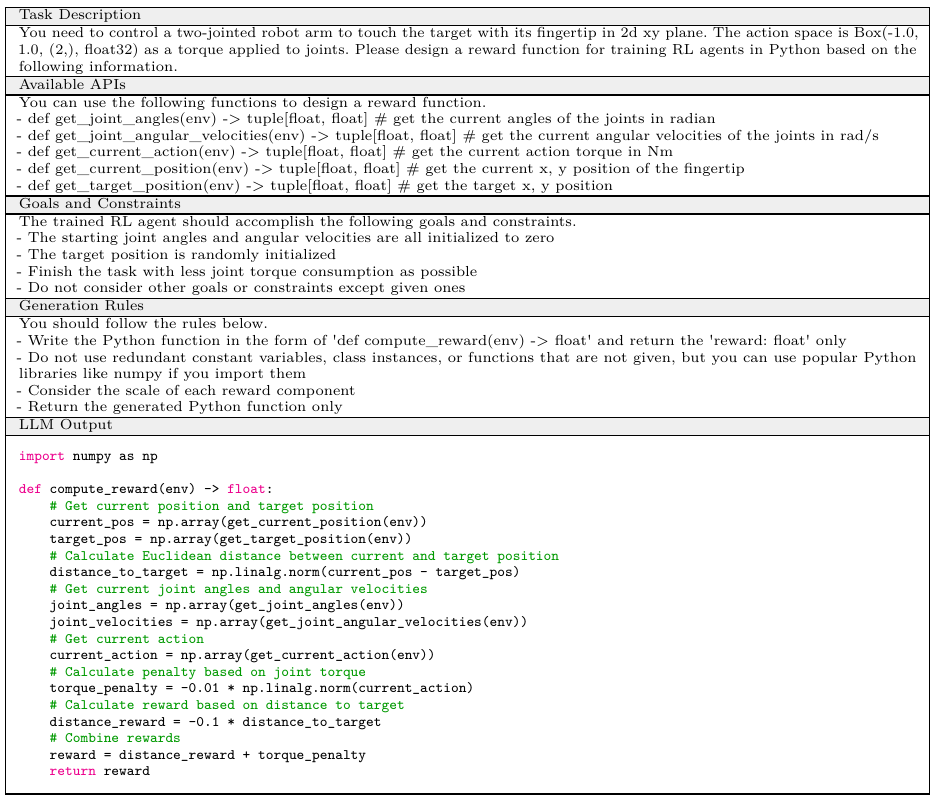}
    \caption{Prompts example for reward function generation. The prompt consists of \textit{task description}, \textit{available APIs}, \textit{goals and constraints}, and \textit{generation rules}. The LLM generates a reward function in Python code for RL training.}
    \label{tab:prompt_reward}
\end{figure*}
\noindent

\subsection{Planning Prompt: Few-shot Planning}
We present a planning prompt design aimed at predicting subsequent actions to fulfill an instructed objective, integrating contextual elements such as available actions and environmental settings. This design particularly focuses on few-shot planning, enhancing performance through examples. The design comprises four components: (1) \textit{task description}, (2) \textit{examples}, (3) \textit{task context}, (4) and \textit{additional interactions}, detailed in Fig.~\ref{tab:prompt_planning}.

The \textit{task description} include task objectives, expected behaviors, and response formats similar to conventional prompts. However, unlike previous ones, this prompt specifies the robot's constraints, including initial states and action limitations. For example, the term `CANNOT' in Fig.~\ref{tab:prompt_planning} emphasizes the robot's limitation to manipulating only one object per action. Moreover, these constraints extend to the rules governing the `Done' action, indicating task completion.

The \textit{examples} demonstrate input-output pairs that guide the LLM in generating the desired action. The \textit{examples} adapt the generic `object' argument in the allowed actions (e.g., `Close (object)') to specific object names such as `drawer' or `paper,' reinforcing task constraints written in the \textit{task description}. For instance, the second example returns the `Done' signal instead of further planning after achieving the task objective. 

The \textit{task context} provides information about the current scenario, including `task,' `allowed actions,' `visible objects,' `executed plans,' and `next plan,' as shown in the examples. We allow the LLM to fill in the blank space after `next plan:' suggesting the next action without adding unnecessary elements like line breaks, ensuring output precision.

Furthermore, when additional prompts update the executed plans, the LLM generates new plans based on this updated context without reiterating the full task context, enabling a dynamic and iterative planning process that adapts to changes and maintains efficiency.

\subsection{Code-Generation Prompt: Reward Design}
We introduce a code-generation prompt design to generate a reward function for the MuJoCo-based Reacher task~\cite{todorov2012mujoco} from Gymnasium~\cite{towers_gymnasium_2023}. The goal of the Reacher task is to move the end-effector of a robotic arm close to a designated target position from an arbitrary starting configuration. The prompt is to translate this task objective into a reward-specifying code. Fig.~\ref{tab:prompt_reward} shows the design detail, comprising four key elements: (1) \textit{task description}, (2) \textit{available APIs}, (3) \textit{goals and constraints}, and (4) \textit{generation rules}.

\textit{Task description} define the expected robot behavior and task conditions for the LLM, including the robot's control strategies and the action space of the two-joint robot arm. We particularly specify the action space as a continuous `Box' space using an API from Gymnasium, assuming the LLM's familiarity with well-known library functions. Then, this description leads the LLM to grasp the overarching RL objective of the defined actions.

\textit{Available APIs} list the APIs necessary for designing the reward function, including the names and input-output specifications of each API. By providing Python function annotations, we enable the LLM to infer the types of inputs and outputs, given its presumed knowledge of float-like variable types and how the APIs work.

\textit{Goals and constraints} provide the task objectives and limitations that guide the reward contents. We clearly define the initial setup, goal assignment, and goal conditions, aiming to exclude unnecessary reward components, such as penalizing high velocities for smooth motion. Note that we recommend the use of concise and consistent words, such as `torque,' as used in the \textit{task description}, instead of `power.' This ensures the generated reward function aligns with the specified task requirements without introducing ambiguities or unintended penalties.

Lastly, \textit{generation rules} establish guidelines for generating directly executable code, addressing the tendency of LLMs to produce unnecessary or incorrect variables or functions. These rules restrict such declarations, as written in the second component of the \textit{generation rules} in Fig.~\ref{tab:prompt_reward}, encouraging the use of well-known Python libraries to enhance programming quality. Furthermore, considering the linearly combined elements of the reward function, we introduce a rule for scaling reward components to maintain balance.

\section{Conclusion}\label{sec_conc}

In this survey, we have investigated the current robotics research works with large language models in terms of intelligent robot components encompassing communication, perception, planning, and control. This component-wise investigation reveals how researchers integrate LLMs to overcome challenges inherent in pre-LLM approaches across various tasks, thereby offering a comprehensive understanding of LLMs' impact in this field. Within each component area, we examine the improvement of methodologies proposed to maximize the utilization of LLMs’ capabilities and enhance the integrity of their responses.
Additionally, our survey offers guidelines for prompt engineering in each component area, supplemented with key examples of prompt components, to provide practical insights for researcher entering this field.
The core contribution of this paper is to highlight the transformative impact of LLMs in robotics, enabling the development of versatile and intelligent robots.
By synthesizing these insights, we aim to guide future research on integrating LLMs into robotic system.

\bmhead{Acknowledgements}
This research was supported by the DRB-KAIST SketchTheFuture Research Center and the KAIST Convergence Research Institute Operation Program.

\bibliographystyle{sn-basic}
\bibliography{main}

\end{document}